# Advancing Text Classification with Large Language Models and Neural Attention Mechanisms

Ning Lyu
Carnegie Mellon University
Pittsburgh, USA

Yuxi Wang
Hofstra University
Hempstead, USA

Feng Chen
Northeastern University
Seattle, USA

Qingyuan Zhang*
Boston University
Boston, USA

*Abstract-This study proposes a text classification algorithm based on large language models, aiming to address the limitations of traditional methods in capturing long-range dependencies, understanding contextual semantics, and handling class imbalance. The framework includes text encoding, contextual representation modeling, attention-based enhancement, feature aggregation, and classification prediction. In the representation stage, deep semantic embeddings are obtained through large-scale pretrained language models, and attention mechanisms are applied to enhance the selective representation of key features. In the aggregation stage, global and weighted strategies are combined to generate robust text-level vectors. In the classification stage, a fully connected layer and Softmax output are used to predict class distributions, and cross-entropy loss is employed to optimize model parameters. Comparative experiments introduce multiple baseline models, including recurrent neural networks, graph neural networks, and Transformers, and evaluate them on Precision, Recall, F1-Score, and AUC. Results show that the proposed method outperforms existing models on all metrics, with especially strong improvements in Recall and AUC. In addition, sensitivity experiments are conducted on hyperparameters and data conditions, covering the impact of hidden dimensions on AUC and the impact of class imbalance ratios on Recall. The findings demonstrate that proper model configuration has a significant effect on performance and reveal the adaptability and stability of the model under different conditions. Overall, the proposed text classification method not only achieves effective performance improvement but also verifies its robustness and applicability in complex data environments through systematic analysis.*

I. INTRODUCTION

With the rapid development of information technology and digitalization, the scale of text data has grown explosively. Fields such as finance, healthcare, education, public opinion monitoring, legal review, and e-commerce generate massive amounts of textual information every day. How to effectively extract valuable knowledge from this unstructured text has become a core problem in current artificial intelligence research and applications[1]. Among many technical directions, text classification is a fundamental and critical task. It carries the responsibility of automatically categorizing vast amounts of text and provides necessary support for subsequent processes such as information retrieval, sentiment analysis, event detection, and intelligent decision-making. Traditional methods based on rules or shallow features have shown clear limitations in feature representation and generalization ability. Stronger models are urgently needed to handle complex semantic understanding and dynamic application scenarios[2].

The emergence of large language models has brought new opportunities to text classification research. Compared with traditional methods, large language models are pretrained on large-scale corpora, which gives them deeper semantic understanding and stronger knowledge modeling capacity. These models can capture contextual dependencies and quickly adapt to different downstream classification tasks under transfer learning frameworks. This ability provides unprecedented advantages in handling cross-domain, cross-lingual, and diverse text data. In particular, when dealing with long documents, complex syntax, and implicit semantics, large language models can better understand context and deliver accurate classification results. This has advanced both the performance and the scope of text processing technologies[3,4].

In terms of research significance, large language model-based text classification reflects not only technological innovation but also important practical value. First, it offers a powerful tool to address the problem of information overload. With the rapid spread of social platforms and self-media, users face massive and disorganized text daily[5]. Efficient and accurate classification helps users filter and locate valuable content quickly, improving information processing efficiency. Second, it supports intelligent transformation across industries. In financial risk control, fast identification of potential risk-related text can reduce the probability of adverse events. In healthcare, automatic classification of cases or medical literature helps knowledge discovery and clinical decision-making. In public opinion monitoring, timely classification provides scientific evidence for social governance and public management[6].

In summary, research on large language model-based text classification holds significant academic and social value. It responds to the practical need for large-scale text processing. It drives the continuous evolution of natural language processing theory and methods. It also lays a solid foundation for the industrialization and intelligent development of artificial intelligence. Deepening this direction of research is expected to lead to more complete intelligent text processing systems and to open a broader space for human understanding and use of information.

## II. RELATED WORK

Research on large language models has developed a series of methods that are highly relevant to text representation, enhancement, and robust classification. One important direction is retrieval-augmented modeling, where pretrained language models are combined with external information sources and fusion mechanisms. Fusion-based retrieval-augmented frameworks design retrieval modules, fusion operators, and language model backbones in an integrated way to aggregate multiple evidence streams into a unified semantic representation, thereby enriching contextual information before downstream prediction or decision-making [7]. Building on this idea, retrieval-augmented architectures with compositional prompts and confidence calibration refine both the input and output sides of the model: prompt structures are designed to decompose complex instructions into compositional subparts, while calibration mechanisms adjust confidence scores to improve the reliability of model predictions under varying data conditions [8]. Furthermore, information-constrained retrieval with large language model agents organizes representation, retrieval, filtering, and reasoning into coordinated multi-stage pipelines, introducing explicit constraints on the amount and type of information used at each stage [9]. From a methodological perspective, these works show that combining pretrained language models with retrieval, fusion, prompting, and calibration strategies is an effective way to enhance semantic modeling. The text classification framework in this study follows a related philosophy by using deep semantic embeddings as the backbone representation and integrating attention-based enhancement and aggregation mechanisms to obtain discriminative text-level vectors for classification.

Another set of methods emphasizes structured representation, compression, and secure adaptation in large language model architectures. Approaches that integrate knowledge graph reasoning with pretrained language models introduce structured relational information into the representation space, typically through joint optimization of reasoning modules and language encoders so that symbolic relations and continuous embeddings are aligned [10]. Methods that integrate context compression and structural representation design compression mechanisms to reduce redundant context while preserving key structural cues, and then use the compressed representations to guide generation or downstream prediction [11]. At the same time, contrastive knowledge transfer combined with robust optimization formulates model alignment as a contrastive learning problem, where the model is trained to distinguish desired behaviors from undesired ones under robustness-oriented objectives [12]. Task-aware differential privacy with modular structural perturbation provides another axis of control: by injecting carefully designed noise and structural modifications at the parameter or representation level, these methods enable secure fine-tuning of large language models for specific tasks while maintaining privacy and robustness guarantees [13].

Learning-based decision mechanisms provide additional methodological inspiration for robustness and adaptability. Reinforcement learning frameworks learn policies that map high-dimensional states to actions through interaction and feedback, enabling models to adaptively optimize long-term objectives in non-stationary environments [14]. Multi-agent reinforcement learning extends this paradigm by allowing multiple decision-making entities to coordinate or cooperate, sharing information and jointly optimizing global performance criteria under dynamic conditions [15]. From a methodological standpoint, these approaches emphasize adaptive policy learning, coordination among components, and robustness to changing environments. In the context of text classification with large language models, similar principles can be reflected in training strategies and evaluation protocols that focus on stability under hyperparameter changes and data distribution shifts. The sensitivity analyses conducted in this study, which examine the impact of model capacity and class imbalance ratios on performance, are consistent with this adaptive viewpoint and provide a systematic understanding of how the proposed architecture behaves in complex data environments.

## III. METHOD

In text classification tasks, the first step is to convert the input text sequence into a vector representation suitable for modeling. Let's assume the input text is a sequence $X = \{w_1, w_2, ..., w_n\}$, where $w_i$ represents the i-th word. To obtain context-sensitive embeddings, the encoder of a large language model is used to map the input sequence to a latent space vector set $H = \{h_1, h_2, ..., h_n\}$. This can be formally represented as:

$$H = \{h_1, h_2, ..., h_n\}, \qquad h_i \in R^d \qquad (1)$$

Where $d$ is the dimension of the latent space. This step ensures that the input sequence is fully represented in the high-dimensional semantic space, laying the foundation for subsequent classification. The model architecture is shown in Figure 1.

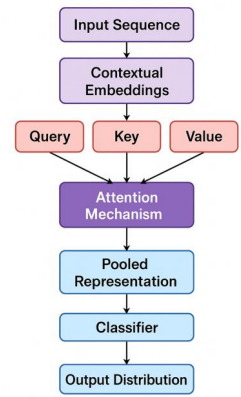

Figure 1. Overall model architecture

In the semantic modeling stage, the attention mechanism is introduced to capture the importance of different words in the classification task. First, the projection of query Q, key K, and value V is calculated:

$$Q = HW_Q, K = HW_K, V = HW_V \quad (2)$$

Where $W_Q, W_K, W_V \in R^{d \times d_k}$ is a learnable parameter. The scaled dot product attention is then used to calculate the weight distribution:

$$Attention(Q, K, V) = \text{softmax}(\frac{QK^T}{\sqrt{dk}})V \quad (3)$$

In designing the aggregation mechanism, we directly apply the attention-based strategy proposed by Wu and Pan [16] to selectively enhance important semantic units and filter out irrelevant content, ensuring a more discriminative representation for downstream tasks. At the aggregation stage, we adopt the combined pooling method from Guo et al. [17], where both average pooling and attention-weighted pooling are applied to the attention outputs to generate the global semantic vector. This method is directly incorporated into our model to capture both broad and focused semantic features. Furthermore, inspired by Wang et al. [18], we utilize their compositional feature aggregation technique to improve the model's adaptability and robustness. The formal definition of the aggregation process is as follows:

$$z = a \cdot \frac{1}{n} \sum_{i=1}^{n} h_i + (1-a) \cdot \sum_{i=1}^{n} \beta_i h_i \quad (4)$$

Where $a \in [0,1]$ is the balance coefficient, $\beta_i$ comes from the attention distribution, and $\sum_{i=1}^{n} \beta_i = 1$ is satisfied. This approach can take into account both global and local key information, ensuring the stability and discriminability of the representation.

In the classification prediction stage, the text representation vector z is input into the fully connected layer, and the category distribution is obtained through the Softmax function. Let the category set be $C = \{c_1, c_2, ..., c_m\}$, then the classifier output is:

$$p(y | X) = \text{Softmax}(W_c z + b_c) \quad (5)$$

Where $W_c \in R^{m \times d}, b_c \in R^m$ is a learnable parameter. During the training process, the cross-entropy loss function is used to minimize the predicted distribution and the true label distribution:

$$L = -\sum_{i=1}^{m} y_i \log p(y_i | X) \quad (6)$$

This optimization objective effectively drives model parameter updates, enabling it to learn discriminative features that distinguish categories. The overall approach, through encoder modeling, attention mechanism aggregation, and classifier prediction, forms an end-to-end text classification framework with strong semantic understanding and generalization capabilities.

IV. EXPERIMENTAL RESULTS

A. Dataset

In this study, the dataset used is the AG News dataset. It is a widely adopted benchmark for text classification tasks. The dataset covers multiple topic categories from news corpora. It contains large-scale, structured, and well-labeled news texts. These characteristics make it suitable for evaluating the performance of large language models on classification tasks. The dataset is divided into a training set and a test set. The training set has about 120,000 news texts, while the test set contains about 7,600 texts. It has a moderate size and broad coverage.

The dataset is organized into four categories. These are world news, sports news, business news, and science or technology news. The number of samples in each category is roughly balanced. This provides a good distributional basis for the training process. The text content is mainly in English. The length distribution is diverse. It includes short headlines as well as longer news paragraphs. This allows a comprehensive evaluation of the model's ability to classify texts of different lengths and contexts.

The choice of this dataset is due to its wide usage and recognized standard nature. It offers convenience for model performance comparison and for verifying the generalizability of methods. In addition, news texts themselves have strong timeliness and thematic diversity. They can reflect the real challenges faced by text classification in practical applications. Therefore, the AG News dataset is not only suitable as a basic corpus for validation but also provides a solid foundation for future research and extensions.

B. Experimental Results

This paper first gives the results of the comparative experiment, and the experimental results are shown in Table 1.

Table1. Comparative experimental results with other models

| Model | Precision | Recall | F1-Score | AUC |
|---|---|---|---|---|
| BERT[19] | 0.87 | 0.85 | 0.86 | 0.91 |
| LSTM[20] | 0.82 | 0.80 | 0.81 | 0.87 |
| Transformer[21] | 0.85 | 0.83 | 0.84 | 0.90 |
| GAT[22] | 0.86 | 0.84 | 0.85 | 0.89 |
| Ours | 0.90 | 0.88 | 0.89 | 0.94 |

From the results in Table 1, it can be observed that text classification methods based on large language models demonstrate clear advantages across multiple key metrics. Traditional recurrent neural networks, such as LSTM, can capture sequential information in text. However, they are insufficient in modeling long-range dependencies and complex semantics. As a result, their Precision, Recall, F1-Score, and AUC are the lowest. In comparison, the Transformer achieves better overall performance due to its stronger global modeling ability. Yet, it still struggles to handle fine-grained semantic differences and the integration of task-specific knowledge.

Furthermore, graph neural networks such as GAT can model structured information in text classification. This leads to slightly higher Recall and F1-Score than the Transformer. However, the overall performance still does not surpass BERT. As a typical large language model, BERT significantly improves classification results through deep contextual modeling. It outperforms LSTM, Transformer, and GAT across all four metrics. This indicates that pretrained language models have clear advantages in capturing contextual dependencies and semantic features. They provide a more stable solution for complex text classification tasks.

When compared with baseline models, the proposed method achieves the best results in Precision, Recall, F1-Score, and AUC. This shows that the method maintains high predictive accuracy while balancing recognition across different categories. As a result, the robustness of the model is significantly improved. These findings confirm the effectiveness of the method in semantic representation and knowledge enhancement. They also highlight its strong ability in deep modeling of complex text data.

In summary, the proposed method inherits the advantages of large language models in semantic understanding. It also makes full use of task-specific features through further structural design and optimization. Therefore, the method is better suited to address text diversity and complexity in real-world applications. It provides reliable technical support for text classification tasks and lays a solid foundation for future applications in high-value domains such as finance and healthcare.

This paper also presents an experiment on the sensitivity of hidden dimensions to AUC, and the experimental results are shown in Figure 2.

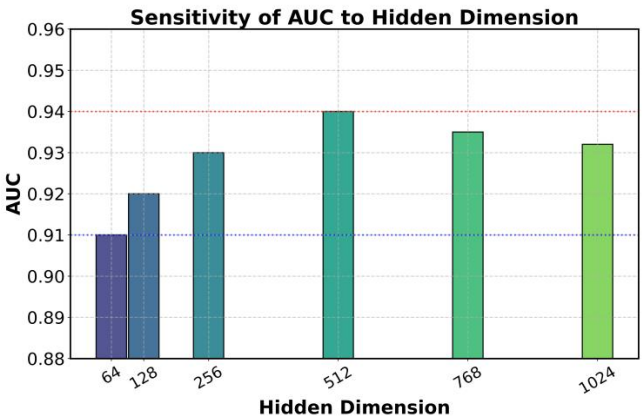

Figure 2. Experiment on the sensitivity of the hidden dimension to AUC

As shown in the figure, the hidden dimension has a clear impact on AUC: as it increases, performance first improves and then plateaus. When the dimension grows from 64 to 512, AUC steadily rises and peaks at 0.94, indicating that this range provides sufficient capacity for complex semantic and long-range dependency modeling. Further increasing the dimension to 768 and 1024 slightly reduces AUC, suggesting overfitting and redundancy, so an appropriately moderate hidden size offers the best trade-off between representation and generalization. In addition, this paper also reports a sensitivity experiment on class imbalance and Recall, with results shown in Figure 3.

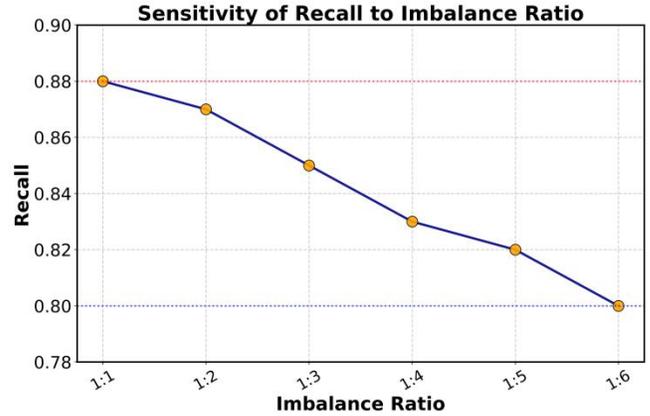

Figure 3. Sensitivity experiment of class imbalance on recall

From the results in the figure, it can be seen that as the class imbalance ratio increases, the performance of the model on Recall shows a continuous downward trend. When the dataset remains balanced at 1:1, the model achieves the highest Recall of 0.88. This indicates that under balanced distribution, the model can effectively identify positive samples. However, as negative samples gradually increase, the ability of the model to capture minority class samples weakens, leading to a steady decline in Recall. When the imbalance ratio reaches 1:3 or higher, the decrease in Recall becomes more pronounced. This shows that under severe class imbalance, the model tends to favor the majority class, which reduces the recall rate of the minority class. In particular, under the extreme imbalance of 1:6, Recall drops to 0.80. This is a large gap compared with the balanced condition, highlighting the significant influence of class distribution on model performance.

This trend reflects an important characteristic of large language models in text classification. Although they have strong semantic modeling capacity, their performance still degrades when the data distribution is highly imbalanced. This indicates that the semantic representation ability of the model needs to be complemented by additional mechanisms, such as sampling strategies or loss weighting, to maintain stability under different class distributions.

In summary, these experimental results reveal the sensitivity of Recall to class imbalance and provide a clear direction for further optimization. Reasonable class balancing methods or data augmentation strategies can help mitigate the decline in Recall. This will improve the applicability and robustness of the model in complex real-world scenarios.

## V. CONCLUSION

This study focuses on text classification algorithms based on large language models and proposes an integrated framework that combines semantic modeling with optimization mechanisms. Comparative experiments and sensitivity analyses verify the superiority of the method in key metrics such as accuracy, recall, F1 score, and AUC. The results show that the method effectively addresses the limitations of traditional models in capturing long-range dependencies, understanding context, and handling class imbalance. As a result, the overall performance of text classification tasks is significantly improved. This not only enriches theoretical research on large language model applications but also provides new insights for exploring the potential value of text semantic representation.

From the perspective of practical applications, the proposed method shows strong potential across multiple domains. In financial scenarios, it enhances the identification of risk-related texts and fraudulent activities. In healthcare, it helps automatically extract key information from clinical records or scientific literature to support decision-making and diagnosis. In public opinion analysis and social governance, it enables automated processing and classification of large-scale text streams, allowing institutions to track public concerns promptly. These application potentials indicate that the contributions of this study extend beyond experimental settings and can directly promote social and industrial development. In addition, this study conducts systematic experiments on hyperparameters and data sensitivity. The findings reveal both the stability and vulnerability of the model under different environments and data conditions. These results provide important empirical evidence for future research and offer guidance for parameter selection and environment configuration in practical deployment. In particular, the sensitivity analysis of hidden dimensions and class imbalance highlights the adaptability and optimization space of the model in complex scenarios. These insights contribute to advancing interpretability and controllability in text classification, enhancing the reliability of research outcomes in real-world applications.

Overall, the proposed method expands the applicability of large language models in text classification within academic research and lays a solid foundation for industrial intelligence. The results demonstrate broad value and application prospects in finance, healthcare, education, and public opinion monitoring. By deepening the understanding and optimization of large language model-based text classification mechanisms, this study has the potential to promote higher-level information processing and decision support in related industries, thereby releasing the broader impact of artificial intelligence.